\pdfoutput=1

\documentclass[11pt]{article}

\usepackage{arxiv}

\usepackage{times}
\usepackage{latexsym}

\usepackage[T1]{fontenc}

\usepackage[utf8]{inputenc}

\usepackage{microtype}

\usepackage{listings} 
\usepackage{graphicx}
\usepackage{caption}
\usepackage{subcaption} 

\usepackage[utf8]{inputenc}
\usepackage[T1]{fontenc}
\usepackage{hyphenat}
\usepackage{xspace}
\usepackage{amsmath}
\usepackage{amsfonts}
\usepackage{url}
\usepackage{booktabs}
\usepackage{multirow}
\usepackage{makecell}
\usepackage{caption}
\usepackage{minibox}
\usepackage{bbm}
\usepackage{graphicx}
\usepackage{balance}
\usepackage{mathtools}
\usepackage{color}
\usepackage{marvosym}
\usepackage{ifthen}
\usepackage{textcomp}
\usepackage{enumitem}
\usepackage{verbatim}
\usepackage{algorithm}
\usepackage{numprint}
\usepackage{balance}

\usepackage{amsthm}
\theoremstyle{plain}

\newcommand{\chatoDisplayMode}[1]{#1}

\definecolor{MyRed}{rgb}{0.6,0.0,0.0} 
\definecolor{MyBlack}{rgb}{0.1,0.1,0.1} 
\newcommand{\inred}[1]{{\color{MyRed}\sf\textbf{\textsc{#1}}}}
\newcommand{\frameit}[2]{
  \begin{center}
  {\color{MyRed}
  \framebox[.9\columnwidth][l]{
    \begin{minipage}{.85\columnwidth}
    \inred{#1}: {\sf\color{MyBlack}#2}
    \end{minipage}
  }\\
  }
  \end{center}
}

\newcommand{\note}[2][]{\chatoDisplayMode{\def\@tmpsig{#1}\frameit{{\Pointinghand} Note}{#2\ifx \@tmpsig \@empty \else \mbox{ --\em #1}\fi}}}
\newcommand{\todo}[2][]{\chatoDisplayMode{\def\@tmpsig{#1}\frameit{{\Writinghand} To-do}{#2\ifx \@tmpsig \@empty \else \mbox{ --\em #1}\fi}}}

\newcommand{\abbrevStyle}[1]{#1}

\newcommand{\ie}{\abbrevStyle{i.e.}\xspace}

\newcommand{\Secref}[1]{Sec.~\ref{#1}}

\newcommand{\Figref}[1]{Fig.~\ref{#1}}

\newcommand{\xhdr}[1]{\vspace{1.7mm}\noindent{{\bf #1.}}}

\newcommand{\textcite}[1]{\citeauthor{#1} \shortcite{#1}}

\newcommand{\hide}[1]{}

\newcommand{\defeq}{\vcentcolon=}

\makeatletter
\newcommand{\iffont}[2]{\ifthenelse{\equal{\f@family}{#1}}{#2}{}}
\makeatother

  \usepackage{mathptmx}

  \DeclareSymbolFont{greek}{OML}{cmm}{m}{n}
  \DeclareMathSymbol{\alpha}{\mathalpha}{greek}{"0B}
  \DeclareMathSymbol{\beta}{\mathalpha}{greek}{"0C}
  \DeclareMathSymbol{\gamma}{\mathalpha}{greek}{"0D}
  \DeclareMathSymbol{\delta}{\mathalpha}{greek}{"0E}
  \DeclareMathSymbol{\epsilon}{\mathalpha}{greek}{"0F}
  \DeclareMathSymbol{\zeta}{\mathalpha}{greek}{"10}
  \DeclareMathSymbol{\eta}{\mathalpha}{greek}{"11}
  \DeclareMathSymbol{\theta}{\mathalpha}{greek}{"12}
  \DeclareMathSymbol{\iota}{\mathalpha}{greek}{"13}
  \DeclareMathSymbol{\kappa}{\mathalpha}{greek}{"14}
  \DeclareMathSymbol{\lambda}{\mathalpha}{greek}{"15}
  \DeclareMathSymbol{\mu}{\mathalpha}{greek}{"16}
  \DeclareMathSymbol{\nu}{\mathalpha}{greek}{"17}
  \DeclareMathSymbol{\xi}{\mathalpha}{greek}{"18}
  \DeclareMathSymbol{\pi}{\mathalpha}{greek}{"19}
  \DeclareMathSymbol{\rho}{\mathalpha}{greek}{"1A}
  \DeclareMathSymbol{\sigma}{\mathalpha}{greek}{"1B}
  \DeclareMathSymbol{\tau}{\mathalpha}{greek}{"1C}
  \DeclareMathSymbol{\upsilon}{\mathalpha}{greek}{"1D}
  \DeclareMathSymbol{\phi}{\mathalpha}{greek}{"1E}
  \DeclareMathSymbol{\chi}{\mathalpha}{greek}{"1F}
  \DeclareMathSymbol{\psi}{\mathalpha}{greek}{"20}
  \DeclareMathSymbol{\omega}{\mathalpha}{greek}{"21}
  \DeclareMathSymbol{\varepsilon}{\mathalpha}{greek}{"22}
  \DeclareMathSymbol{\vartheta}{\mathalpha}{greek}{"23}
  \DeclareMathSymbol{\varpi}{\mathalpha}{greek}{"24}
  \DeclareMathSymbol{\varrho}{\mathalpha}{greek}{"25}
  \DeclareMathSymbol{\varsigma}{\mathalpha}{greek}{"26}
  \DeclareMathSymbol{\varphi}{\mathalpha}{greek}{"27}
  \DeclareSymbolFont{otone}{OT1}{cmr}{m}{n}
  \DeclareMathSymbol{\Gamma}{\mathalpha}{otone}{0}
  \DeclareMathSymbol{\Delta}{\mathalpha}{otone}{1}
  \DeclareMathSymbol{\Theta}{\mathalpha}{otone}{2}
  \DeclareMathSymbol{\Lambda}{\mathalpha}{otone}{3}
  \DeclareMathSymbol{\Xi}{\mathalpha}{otone}{4}
  \DeclareMathSymbol{\Pi}{\mathalpha}{otone}{5}
  \DeclareMathSymbol{\Sigma}{\mathalpha}{otone}{6}
  \DeclareMathSymbol{\Upsilon}{\mathalpha}{otone}{7}
  \DeclareMathSymbol{\Phi}{\mathalpha}{otone}{8}
  \DeclareMathSymbol{\Psi}{\mathalpha}{otone}{9}
  \DeclareMathSymbol{\Omega}{\mathalpha}{otone}{10}
  \DeclareSymbolFont{syms}{OML}{cmm}{m}{it}
  \DeclareMathSymbol{\partial}{\mathord}{syms}{"40}
  \DeclareMathAlphabet{\mathbold}{OML}{cmm}{b}{it}
  \DeclareSymbolFont{largesymbols}{OMX}{cmex}{m}{n}

\usepackage{algpseudocode} 
\usepackage{placeins}
\usepackage{float}
\usepackage{tabularx}
\usepackage[dvipsnames]{xcolor}	
\definecolor{darkblue}{rgb}{0, 0, 0.5}
\usepackage{wrapfig}

\usepackage[most]{tcolorbox}

\usepackage[colorlinks=true, citecolor=darkblue, linkcolor=darkblue, urlcolor=darkblue]{hyperref}


\newcommand{\semtok}[1]{\mathbf{#1}^{\sigma}}

\DeclareSymbolFont{extraup}{U}{zavm}{m}{n}
\DeclareMathSymbol{\epfl}{\mathalpha}{extraup}{81}
\DeclareMathSymbol{\cnrs}{\mathalpha}{extraup}{83}

\definecolor{myblue}{RGB}{0, 102, 204}

\newtcolorbox{summarybox}{
  colback=myblue!5!white,
  colframe=myblue!75!black,
  coltext=myblue!75!black,
  sharp corners,
  rounded corners=northeast,
  arc is angular,
  arc=3mm,
  top=6pt,
  bottom=6pt,
  left=6pt,
  right=6pt,
  boxsep=8pt,
  boxrule=1pt,
}

\title{Agentic AI: The Era of Semantic Decoding}

\author{
Maxime Peyrard,$^{\cnrs\ast}$
Martin Josifoski,$^{\epfl\ast}$
Robert West$^{\epfl}$ \\
    $^{\cnrs}$Univ. Grenoble Alpes, CNRS, Grenoble INP, LIG \quad $^{\epfl}$EPFL\\
    {maxime.peyrard@univ-grenoble-alpes.fr} \\
    {\{martin.josifoski, robert.west\}@epfl.ch} \\
}

\begin{document}
\maketitle

\renewcommand*{\thefootnote}{\fnsymbol{footnote}}
\footnotetext[1]{Equal contribution.}

\begin{abstract}
Recent work demonstrated great promise in the idea of orchestrating collaborations between large language models (LLMs), human input, and various tools to address the inherent limitations of LLMs. 
We propose a novel perspective called \textit{semantic decoding}, which frames these collaborative processes as optimization procedures in semantic space. Specifically, we conceptualize LLMs as semantic processors that manipulate meaningful pieces of information that we call semantic tokens (also known as thoughts). LLMs are among a large pool of other semantic processors, including humans and tools, such as search engines or code executors. Collectively, semantic processors engage in dynamic exchanges of semantic tokens to progressively construct high-utility outputs. We refer to these orchestrated interactions among semantic processors, optimizing and searching in semantic space, as \textit{semantic decoding algorithms}.
This concept draws a direct parallel to the well-studied problem of syntactic decoding, which involves crafting algorithms to best exploit auto-regressive language models for extracting high-utility sequences of syntactic tokens. By focusing on the semantic level and disregarding syntactic details, we gain a fresh perspective on the engineering of AI systems, enabling us to imagine systems with much greater complexity and capabilities.
In this position paper, we formalize the transition from syntactic to semantic tokens as well as the analogy between syntactic and semantic decoding. Subsequently, we explore the possibilities of optimizing within the space of semantic tokens via semantic decoding algorithms. We conclude with a list of research opportunities and questions arising from this fresh perspective.
The semantic decoding perspective offers a powerful abstraction for search and optimization directly in the space of meaningful concepts, with semantic tokens as the fundamental units of a new type of computation that we call \textit{pragmatic computing}. We say pragmatic because the optimization of utility via the exchange of semantic tokens is a computation that gives rise to a dynamic and task-dependent notion of meaning.
\end{abstract}

\section{Introduction}
\label{sec:introduction}
Recent research suggests that strategically orchestrated collaborations between large language models (LLMs), tools, and humans can effectively overcome LLMs' inherent limitations, leading to substantial performance improvements \cite{sel2023algorithm, FunSearch2023, ding2023thoughts, yao2023tree, besta2023graph, wang2023selfconsistency, wang2023jarvis1, shinn2023reflexion, dasgupta2023collaborating, du2024anytool}.

To conceptualize this evolution, one can consider LLMs as generators of semantically coherent text fragments, often referred to as \textit{thoughts} or, equivalently in this work, \textit{semantic tokens} \cite{CoT,yao2023tree,besta2023graph,ding2023thoughts,sel2023algorithm}. 
This viewpoint positions LLMs as just another kind of contributor to a diverse pool of what we call \textit{semantic processors}, which includes humans, search engines, external memories, code executors, and more.
Collectively, these semantic processors engage in a dynamic process, exchanging and manipulating semantic tokens to progressively construct a high-utility semantic token as output \cite{ding2023thoughts, josifoski2023flows}. To structure the vast space of possible collaboration strategies, several frameworks have been proposed, such as LangChain \citep{langchain2022}, aiFlows \cite{josifoski2023flows}, MetaGPT \citep{DBLP:journals/corr/abs-2308-00352}, SwarmGPT \cite{jiao2023swarmgpt}, and AutoGen \citep{DBLP:journals/corr/abs-2308-08155}, among others.

\begin{wrapfigure}{r}{0.5\textwidth}
    \centering
    \includegraphics[width=0.85\linewidth]{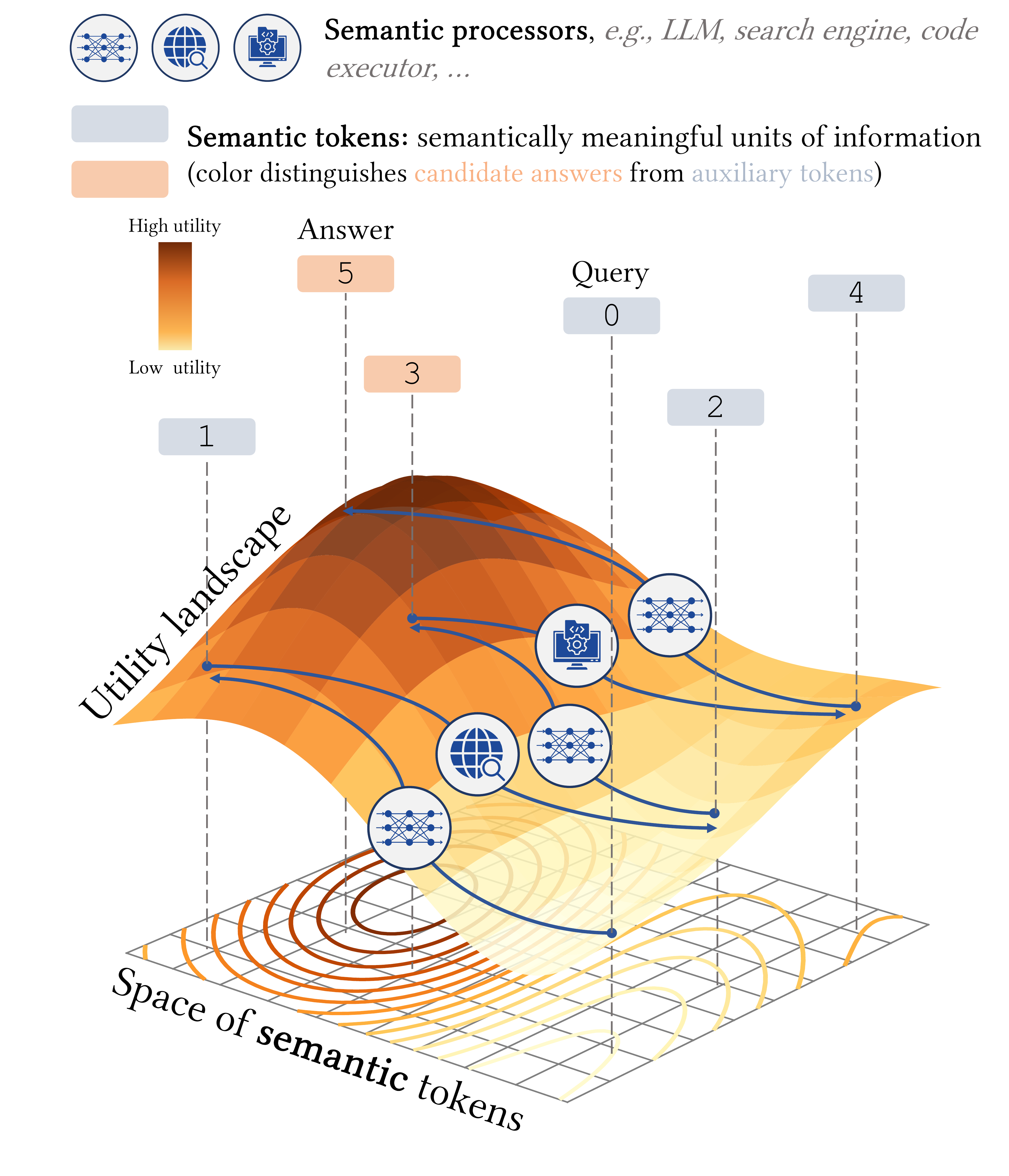}
    \caption{\textbf{Illustration of semantic decoding: optimizing utility in the space of semantic tokens.} 
    Semantic tokens -- semantically coherent units of text -- form the basic units of communication among what we call \textbf{semantic processors}, which includes LLMs, humans, and various tools. Then, \textbf{utility} is a function defined over the space of semantic tokens, indicating a semantic token (or stream of tokens) solves the task. A \textbf{semantic decoding algorithm} orchestrates the exchange of semantic tokens among semantic processors to robustly extract a high-utility semantic token. This orchestration can be viewed as a search and optimization procedure within the semantic space. Throughout the decoding process, auxiliary tokens (depicted in gray) are generated; while these tokens are not answers themselves and have low utility, they serve as anchor points for further exploration toward regions of higher utility. Examples of auxiliary tokens include feedback or grounding information. The generation of auxiliary tokens should increase the expected utility of the trajectory in the semantic space. This example is a simplified illustration of a basic trajectory in the semantic space, we show in \Secref{sec:sem_decode} and \Secref{sec:research_opportunities} that semantic decoding can be used in much more complex and creative ways.}
    \label{fig:fig1}
\end{wrapfigure}

This work presents a different kind of perspective on the advancements in AI collaboration, independent of, but consistent with, these frameworks. Rather than proposing an abstract model of communication between semantic processors, we focus on the optimization that the interaction is globally performing in semantic space to search for the solution. 
We call this perspective \textit{semantic decoding} because it views semantic tokens as the basic units of a new type of computation happening directly in the space of semantic tokens. Then, a \textit{semantic decoding algorithm} is an orchestrated interaction between semantic processors that performs optimization and search in semantic space to reliably construct high-utility trajectories. Our perspective is visually summarized in \Figref{fig:fig1}.

This optimization perspective draws a direct analogy with the well-known problem of \emph{syntactic decoding}, where an algorithmic process -- the decoding algorithm -- aims to extract a high-utility sequence of (syntactic) tokens while being guided by the next token distribution of an auto-regressive language model. 
To build around the inherent limitations of auto-regressive language models, the field of syntactic decoding produced a rich set of techniques leveraging external information and heuristics to guide the search in the space of token sequences, optimizing for utility \cite{meister-etal-2020-beam,meister-etal-2023-efficacy,josifoski-etal-2023-language,pmlr-v97-kool19a, NIPS2017_2b24d495,  krishna-etal-2022-rankgen,chaffin-etal-2022-ppl}.

The shift from syntactic to semantic decoding may, at first glance, appear trivial because, clearly, a semantic token is just a sequence of syntactic tokens that language models were anyway designed to produce. However, by abstracting away from the syntactic details and focusing on the computation in the semantic space, a fresh perspective on the engineering of AI systems and their capabilities emerges. This opens the door to imagining systems with much greater complexity. To highlight the importance of good abstractions for innovation, note that the development of modern LLMs would have been infeasible if one had to program them directly in terms of sequences of bits. Not only would this task be overwhelmingly complex for humans to do, but without the proper mental abstractions, it would have been impossible to even imagine something as complex as an LLM. In our view, the field is now ready to move to yet another powerful abstraction where semantic, rather than syntactic, tokens become the building blocks of a new type of computation, one that operates directly in the space of meaningful thoughts and concepts.

In this article, we begin by articulating the transition from manipulating syntactic tokens to manipulating semantic tokens (\Secref{sec:tokens}). Notably, we underscore that the combination of a language model with a syntactic decoding algorithm creates an engine capable of interpreting semantic tokens as input and generating semantic tokens as output, effectively transforming it into a semantic processor.

Moving on to \Secref{sec:decoding}, we draw the analogy between syntactic and semantic decoding. Specifically, we introduce a generalized notion of a decoding algorithm as an algorithmic layer that operates on top of token processors. This layer orchestrates search and optimization over tokens to robustly extract high-utility outputs. We argue that semantic decoding is a pragmatic computation because the optimization of utility via the exchange of semantic tokens is a computation that gives rise to a dynamic and task-dependent notion of meaning.

In \Secref{sec:sem_decode}, our focus shifts to semantic decoding algorithms, where we categorize the types of optimization performed in semantic space into three distinct groups:
\begin{itemize}
    \item 
\textbf{Grammars of thoughts}, encompassing fixed heuristic patterns such as chain-of-thoughts \cite{CoT}, planning before implementing \cite{wang2023describe, wang2023jarvis1, gao2023pal, paul2023refiner, josifoski2023flows}, or relying on fixed feedback or grounding mechanisms \cite{weng2023large, paul2023refiner, josifoski2023flows}. These approaches can be loosely perceived as the semantic-level generalization of \textit{grammar-constraint decoding} at the syntactic level \cite{tromble-eisner-2006-fast, scholak-etal-2021-picard, roy2022benchclamp, geng-etal-2023-grammar}. 
    \item 
\textbf{Guided search}, representing methods that sample and search the semantic space while being guided by \textit{value models}. 
Noteworthy examples include Tree-of-Thought \cite{yao2023tree} and FunSearch \cite{FunSearch2023}, both utilizing large language models (LLMs) to sample semantic tokens and guiding the overall decoding process with value models. 
This category is the semantic equivalent of \textit{value-guided beam search} (VGBS) \cite{NIPS2017_2b24d495, DBLP:conf/cvpr/RenWZLL17, krishna-etal-2022-rankgen} and \textit{Monte Carlo tree search} (MCTS) variants \cite{chaffin-etal-2022-ppl, josifoski-etal-2023-language}. 
    \item 
\textbf{Learning to optimize}, comprising methods that fully embrace the optimization perspective by learning effective ways to navigate the semantic space. Currently underexplored, this category covers methods in which individual components are trained or fine-tuned to be better collaborators, or controllers to route semantic tokens to appropriate semantic processors at the correct time step. These represent the semantic-level counterparts of the  paradigm of \textit{learning to decode} at the syntactic level \cite{wiseman-rush-2016-sequence, pmlr-v97-collobert19a}.
\end{itemize}

In \Secref{sec:research_opportunities}, we present an extensive, albeit non-exhaustive, list of research opportunities and questions emerging from the semantic decoding perspective. This encompasses topics such as (meta-)prompt engineering, learning to optimize in semantic space, synthetic data flows, human--computer interactions, evaluation, interpretability, control, ethics of semantic decoding algorithms, as well as the infrastructure necessary to support such developments.

To help the reader, we provide a glossary of key terms in \Secref{sec:glossary}. Additionally, each section concludes with a concise summary highlighting the key points for easy reference.

\section{From Syntactic to Semantic Tokens}
\label{sec:tokens}
At its core, computation is a syntactic process, where fundamental information building blocks are manipulated. 
These building blocks might take the form of binary digits (bits), abstract symbols in some computational models, or tokens in language models. As Claude Shannon himself noted, computation is syntactic because the symbols lack inherent \textit{meaning}  \cite{shannon}; instead, meaning arises externally through context, via the processes that manipulate them, and the outcomes they produce for external actors.

Since Shannon, many have tried to lift the theory of information processing from the syntactic to the semantic level. 
The prevailing idea is to shift to computational models wherein basic symbols inherently carry semantic content, readily understandable, with meaningful impact on the actors outside the computation \cite{6004632, Carnap1954-CARAOO-3, peyrard-2019-simple}. This idea is illustrated by the concept of \textit{semantic units} in the semantic theory of information \cite{IS4SI-2017-04000,floridi}, or the proposal of a \textit{language of thoughts} \cite{Fodor1975-FODTLO}, postulating that thinking operates on atomic units of meaningful content \cite{sep-language-thought}. 

The shift discussed in this article is of a similar kind, moving from language models processing syntactic tokens to interactions between LLMs and tools whose basic units of computation are \textit{semantic tokens}, i.e., concepts intelligible for users outside of the computation. These semantic tokens are not abstract symbols awaiting to be interpreted, but active carriers of meaning, embodying concepts and ideas directly. We now describe formally syntactic and semantic tokens.

\subsection{Syntactic Tokens}
Syntactic tokens serve as the fundamental computational building blocks in modern natural language processing systems. 
The finite collection of all possible tokens forms a \textit{syntactic vocabulary} $\Sigma$.
These syntactic tokens are designed to be assembled into sequences through concatenation. Let $\mathbf{x} \in \Sigma^{*}$ represent one such sequence, defined as $\mathbf{x} \defeq \langle x_0, \dots, x_m  \rangle $.
Typically, syntactic tokens may consist of words or characters, but more commonly, they are sub-word units.
The set of these units is often learned based on the frequency of character co-occurrences in the available training data with methods such as byte-pair encoding (BPE) \cite{bpe_origin, sennrich-etal-2016-neural}. The algorithm that breaks down natural language into a set of tokens and therefore defines the vocabulary $\Sigma$ is called the \textit{tokenizer}.

\subsection{Syntactic Token Processors: Language Models}
Syntactic tokens are symbols, and language models are computational processes that manipulate them. In particular, a probabilistic language model (PLM) induces a probability distribution $P$ over all possible strings, $\Sigma^{*}$, that can be constructed from the vocabulary of syntactic tokens $\Sigma$. 
The purpose of the language model is to read an input sequence $\mathbf{x}$ of tokens and guide the assembling of an output sequence $\mathbf{y}$.

To efficiently represent a probability distribution over the large combinatorial space of all possible strings, language models use an auto-regressive decomposition, meaning that they model the probability of each subsequent token given a sequence of previous tokens: $P(y_t|\mathbf{y}_{<t})$, where $\mathbf{y}_{<t} := \langle y_0, \dots, y_{t-1} \rangle$. Most modern language models are parameterized by a Transformer architecture with trainable weights $\theta$ \cite{NIPS2017_3f5ee243}.
Then, the probability distribution of an output sequence $\mathbf{y} \defeq \langle y_0, \dots, y_n \rangle$, potentially conditioned on an input $\mathbf{x} \defeq \langle x_0, \dots, x_m \rangle$, is given by
\begin{equation}
    P(\mathbf{y} | \mathbf{x}) = \prod\limits_{t=0}^{|\mathbf{y}|} p_{\theta} (y_{t} | \mathbf{y}_{<t}, \mathbf{x}).
\end{equation}
Importantly, the language model alone does not directly tell us how to produce an output sequence; it only specifies a probability distribution over the next token given a partial sequence. As described in \Secref{sec:decoding}, the combination with a decoding algorithm is necessary to transform a language model into a system that can produce output sequences.

\subsection{Semantic Tokens}
We define a \textit{semantic token}, also referred to as a \textit{thought}, as a sequence of syntactic tokens that conveys \textit{meaningful information}. A semantic token, denoted as $\semtok{x} \in \Gamma$, is an element of \textit{semantic vocabulary} $\Gamma$, a subset of $\Sigma^{*}$, representing the potentially infinite set of semantic tokens. It is important to note that not all syntactically valid strings convey meaningful information; thus, $\Gamma \neq \Sigma^{*}$. To differentiate semantic tokens (e.g., $\semtok{x}$) from arbitrary strings (e.g., $\mathbf{x}$), we use the superscript $\sigma$. 

What exactly defines a semantic token? When can we determine that a string is ``semantically meaningful''? 
Semantic tokens are embedded within natural language, drawing their meaning directly from potential human interpretation. Similar to how words derive meaning through contextual usage within a language, observed by external actors, the meaning of a string is shaped by its interaction with other elements in the system and its relationship with other semantic tokens.
Finally, they also carry \textit{pragmatic meaning} due to the effects they induce on the computation. For instance, both the input and output sequences of an AI system are semantic tokens, with meaning assigned through their usage. The input sequence $\semtok{x}$ is \textit{intended to} prompt a specific response or behavior from the AI system, while the output sequence $\semtok{y}$ is \textit{interpreted as} a response to a query, serving a particular purpose. Semantic tokens extend beyond input and output sequences; they encompass any other text conveying meaningful information. This includes thoughts, as defined and utilized in various frameworks such as chain-of-thoughts \cite{CoT}, tree-of-thought \cite{yao2023tree, long2023large, xie2023selfevaluation}, and graph-of-thoughts \cite{besta2023graph, yao2023chainofthought}.

\subsection{Semantic Token Processors}

\xhdr{Probabilistic semantic token model} 
A probabilistic language model is trivially also a \textit{probabilistic semantic token model}. Due to the auto-regressive decomposition, the language model induces a probability over strings and therefore over thoughts: $p_{\theta}(\semtok{y}|\semtok{x})$. In general, if there is a history of previously generated semantic tokens $H = \left[ \semtok{x}_0, \dots, \semtok{x}_m \right]$, then the language model induces a probability distribution on the next semantic token: 
\begin{equation}
    P(\semtok{y}|H) = P(\semtok{y}|\semtok{x}_0, \dots, \semtok{x}_m), 
\end{equation}
which is, under the hood, still decomposed via an auto-regressive model on syntactic tokens:
\begin{equation}
    P(\semtok{y}|H) = \prod\limits_{t=0}^{|\mathbf{y}|} p_{\theta} (y_{t} | \mathbf{y}_{<t}, \semtok{x}_0, \dots, \semtok{x}_m).
\end{equation}
In fact, this property makes the perspective shift from syntactic to semantic token possible, the language model being the enabler of the shift. 

\xhdr{From language models to semantic processors}
A \textit{semantic token processor}, or \textit{semantic processor} in short, is a system designed to take a semantic token $\semtok{x}$ as input and generate a corresponding output semantic token $\semtok{y}$. 
While numerous systems fall under this category, let us focus on a semantic processor based on a language model. 
The language model, by itself, provides only a distribution over the next syntactic tokens. However, to generate a sequence of syntactic tokens, it needs to be paired with a \textit{decoding algorithm}. Essentially, a decoding algorithm is responsible for utilizing the distribution of the next tokens and determining which syntactic tokens to assemble into the output string. In practice, language models are commonly paired with decoding algorithms like top-k \citep{fan-etal-2018-hierarchical}, top-p \citep{DBLP:conf/iclr/HoltzmanBDFC20}, often in combination with beam search. Various decoding heuristics and algorithms have been extensively explored to address the challenges inherent in the auto-regressive nature of language models. The decoding problem is the primary focus of \Secref{sec:decoding}.

It is important to note that employing the same language model with different decoding algorithms yields distinct outputs, thereby constituting different semantic processors. Moreover, other factors such as the prompting scheme can further differentiate semantic processors. For example, the chain-of-thought \cite{CoT} or least-to-most prompting \cite{zhou2023leasttomost} schemes generally result in different outputs, even when applied to the same LLM, thus defining different semantic processors.

\xhdr{Other semantic processors}
While syntactic tokens exist for technical purposes, to support the practical computation of language models, semantic tokens are ubiquitous as they are the preferred units of human thinking.
Semantic tokens are manipulated by a great variety of semantic processors, with humans being the primary example. 
Additionally, the tools humans build serve a purpose: to transform a meaningful input into a meaningful and useful output. Naturally, their inputs and outputs correspond to semantic tokens. Humans leverage tools like search engines, code executors, databases, APIs, etc., directly on a daily basis. 
There are substantial research efforts focused on developing methods for augmenting LLMs with such tools \cite{DBLP:journals/corr/abs-2302-07842}.
This perspective holds significance as AI systems, like any other tool, can now participate in a rich ecosystem of semantic processors that communicate via the exchange of semantic tokens.

\begin{summarybox}
\textbf{Summary of the shift from syntactic to semantic tokens:} \\
At the syntactic level, \textbf{syntactic tokens} are determined by the tokenizer and act as the basic symbols manipulated by language models, the \textbf{syntactic token processors}.

At the semantic level, \textbf{semantic tokens}, or \textbf{thoughts}, are meaningful units of information. 
\textbf{Semantic processors} are the processes that manipulate these semantic tokens. When combined with a decoding algorithm, language models become semantic processors and join the rich ecosystem of semantic processors that includes humans and tools.
\end{summarybox}

\section{Decoding: Extracting Utility from Token Processors}
\label{sec:decoding}

\begin{figure*}[t]
    \centering   
    \includegraphics[scale=1, width=\textwidth]{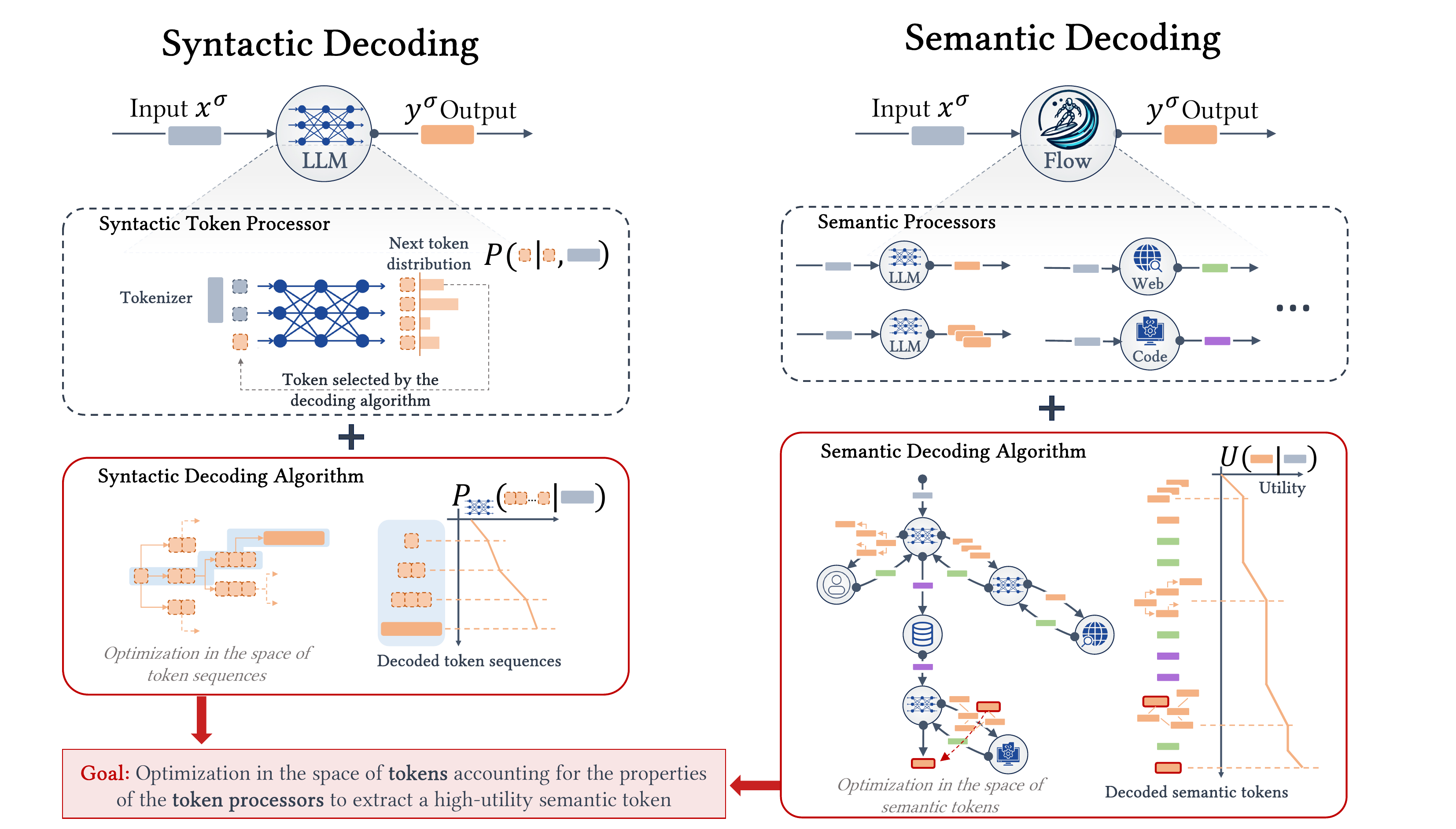}
    \caption{\textbf{Illustrating the analogy between syntactic and semantic decoding perspectives}
    On the left side, we depict the conventional (syntactic) decoding process of generating sequences of tokens from an auto-regressive language model. 
    The decoding algorithm strategically uses the language model to produce a high-utility output sequence.
    Similarly, we frame recent progress in AI, human, and tool collaboration as the semantic-level analogy of this process. 
    Here, the computational units are referred to as \textit{semantic processors}, which manipulate semantic tokens, representing semantically coherent pieces of text. The semantic decoding algorithm harnesses the capabilities of the semantic processors, orchestrating their computation through semantic token exchange to extract a high-utility output.
    Both perspectives share a common objective: extracting high-utility output by leveraging the token processors available during inference. 
}
    \label{fig:syn_sem}
\end{figure*}

In the previous section, we introduce a more general definition of a token and the concept of a token processor -- the components that make up modern AI systems. Now, our focus shifts to the algorithm layer responsible for orchestrating these components to solve practical tasks. We refer to this layer as the \textit{decoding algorithm}
We draw an analogy between decoding operating on syntactic tokens and decoding operating on semantic tokens.
\Figref{fig:syn_sem} visually depicts this analogy.

\subsection{General Formulation of the Decoding Problem}
\xhdr{The objective: maximizing utility}
To solve a given task, an AI system processes an input semantic token (a query) aiming to produce optimal output. 
This is where the concept of a \textit{utility function}, denoted as $u_t(\semtok{y} | \semtok{x})$, becomes critical. 
It scores candidate semantic token outputs, assessing how effectively they solve the task for the specific input $\semtok{x}$. 
For instance, in machine translation, the utility function might judge how well the translation $\semtok{y}$ retains the original meaning conveyed by $\semtok{x}$.
In an ideal scenario, when presented with input $\semtok{x}$, the system selects the output with the highest utility score: $\operatorname*{argmax}_{\semtok{y} \in \Gamma} u_t(\semtok{y} | \semtok{x})$.

In practice, the AI system does not have access to the utility function during inference. 
However, it can rely on value models, which estimate the utility function for partial outputs, to guide the decoding process effectively.

\xhdr{The decoding problem}
AI systems construct their answers by manipulating tokens, the relevant units of computation, using specialized computational tools, the token processors. A dedicated decoding algorithm orchestrates the execution of these basic components to solve the given task, taking into account the properties, capabilities, and limitations of the token processors and organizing the computation to robustly extract high-utility outputs.
At the syntactic level, the syntactic decoding algorithm utilizes syntactic token processors (language models) to manipulate syntactic tokens. By analogy, we propose to conceptualize the orchestrated collaboration of AI, humans, and tools as the semantic equivalent, wherein the semantic decoding algorithm utilizes semantic processors to manipulate semantic tokens. Both algorithms perform optimization within the token space with the objective of extracting high-utility semantic tokens.

\subsection{Syntactic Decoding}
\label{ssec:syn_decode}

The goal of syntactic decoding algorithms is to retrieve an element maximizing a given utility function from the space of the possible token sequences. 
As a primary source of signal, syntactic decoding algorithms rely on the probability distribution induced by a language model and aim to retrieve a sequence that maximizes the model's likelihood. 
However, this approach faces two main challenges. Firstly, the optimization problem $\operatorname*{argmax}{y \in \mathcal{Y}} p\theta(y;|;x)$ becomes intractable due to the exponentially large state space, necessitating approximation techniques. The commonly employed strategies to tackle this difficulty rely on greedy heuristics, such as beam search \citep{sutskever2014sequence}, which focuses on the most probable tokens, either deterministically selecting the top-k candidates or sampling from the top-k or top-p tokens from the distribution \citep{fan-etal-2018-hierarchical}. 
Second, outputs associated with high likelihood are not necessarily of high utility, and effective decoding algorithms need to account for the potential misalignment \citep{josifoski-etal-2023-language}.
Methods to address this problem include (i) value-guided beam search, which uses a greedy strategy similar to beam search but selects the next token using a linear combination of the model's likelihood and the scores from a value model; and (ii) Monte Carlo tree search (MCTS) \cite{chaffin-etal-2022-ppl}, which allocates a fixed computation budget for an informed exploration of multiple paths in the decoding tree before token selection. 
See \citet{josifoski-etal-2023-language} for a more comprehensive overview of prior work on syntactic decoding from an optimization perspective.

\subsection{Semantic Decoding}
\label{ssec:sem_decode}

We now discuss the proposal of \textit{semantic decoding}. 
In this perspective, the objective remains the same as in syntactic decoding: solving a task by generating a high-utility semantic token. However, the fundamental unit of computation shifts from syntactic tokens to semantic ones, and the basic computational processes shift from syntactic processors to semantic ones.

A semantic decoding algorithm coordinates the exchange of semantic tokens among semantic processors to navigate through the semantic token space and identify a high-utility trajectory in semantic space.  

Chain-of-thought (CoT) is a fundamental example, which generates a sequence of thoughts (semantic tokens) before arriving at an answer.  In the initial version, the semantic tokens in the chain do not interact with any other semantic processors or algorithmic components. However, numerous variants have quickly emerged, including:
(i) Sampling and combining multiple reasoning chains \cite{wang2023selfconsistency}, 
(ii) Incorporating feedback from other models, symbolic tools, or human inputs, \cite{wang2023describe, wang2023jarvis1, gao2023pal, paul2023refiner, josifoski2023flows}, 
(iii) Exploring non-linear trajectories in semantic space, such as trees \cite{yao2023tree, long2023large, xie2023selfevaluation} or graphs \cite{besta2023graph, yao2023chainofthought},  and
(iv) Utilizing evolutionary algorithms to select promising semantic tokens,  e.g., FunSearch \cite{FunSearch2023} or PromptBreeder \cite{fernando2023promptbreeder}.

The syntactic decoding setup often operates under the constraint to produce the full trajectory of syntactic tokens as its output. In contrast, semantic decoding has a more flexible ability to manipulate its trajectories and select only a subset of the trajectory as the final output, often selecting its last semantic token as the output.
The constraint to output the entire trajectory makes every decision critical for the quality of the answer. This also renders value estimation more challenging because the value model has to estimate the expected utility for partial outputs that may not look like a candidate's answer. However, at the semantic level, the semantic decoding algorithm has the flexibility to backtrack and remove parts of the trajectory.
It's worth noting that, in principle, nothing prohibits the syntactic decoding algorithm from producing subsets of the trajectory as the output. Recent examples include the usage of pause tokens, which are automatically removed from the final output but used during decoding to allow the model more computational steps.

In \Secref{sec:sem_decode}, we explore how adopting the semantic decoding perspective, which focuses on the optimization and search performed in the semantic space, enables us to broaden our understanding of what is possible for orchestrated interactions.

\subsection{Connections with Pragmatics and Semiotics}
Pragmatics, a subfield of linguistics, focuses on the context-dependent aspects of meaning and how language is used to achieve specific goals \cite{birner2012introduction}. 
Semantic decoding can then be seen as performing pragmatic computation as it relies on the goal-oriented usage of language. Pragmatic computing is an optimization process in semantic space happening via the orchestrated exchange of semantic tokens.
The interaction between semantic processors reflects the pragmatic idea that meaning is constructed through the dynamic exchange of information between goal-driven participants in a conversation.

The semantic decoding perspective also intersects with semiotics \cite{chandler2022semiotics}, the study of signs, symbols, and their interpretation. 
By bridging syntactic and semantic tokens, language models bring meaningless symbols into a space where semantic decoding algorithms can interpret and manipulate these concepts based on their pragmatic usage from various other semantic processors.

By focusing on the optimization of utility through the interaction of semantic processors, the semantic decoding perspective encompasses both the semiotic notion of meaning emergence and the pragmatic emphasis on the context-dependent usage of language in achieving specific goals.

\begin{figure*}[t]
    \centering   
    \includegraphics[scale=1, width=0.86\textwidth]{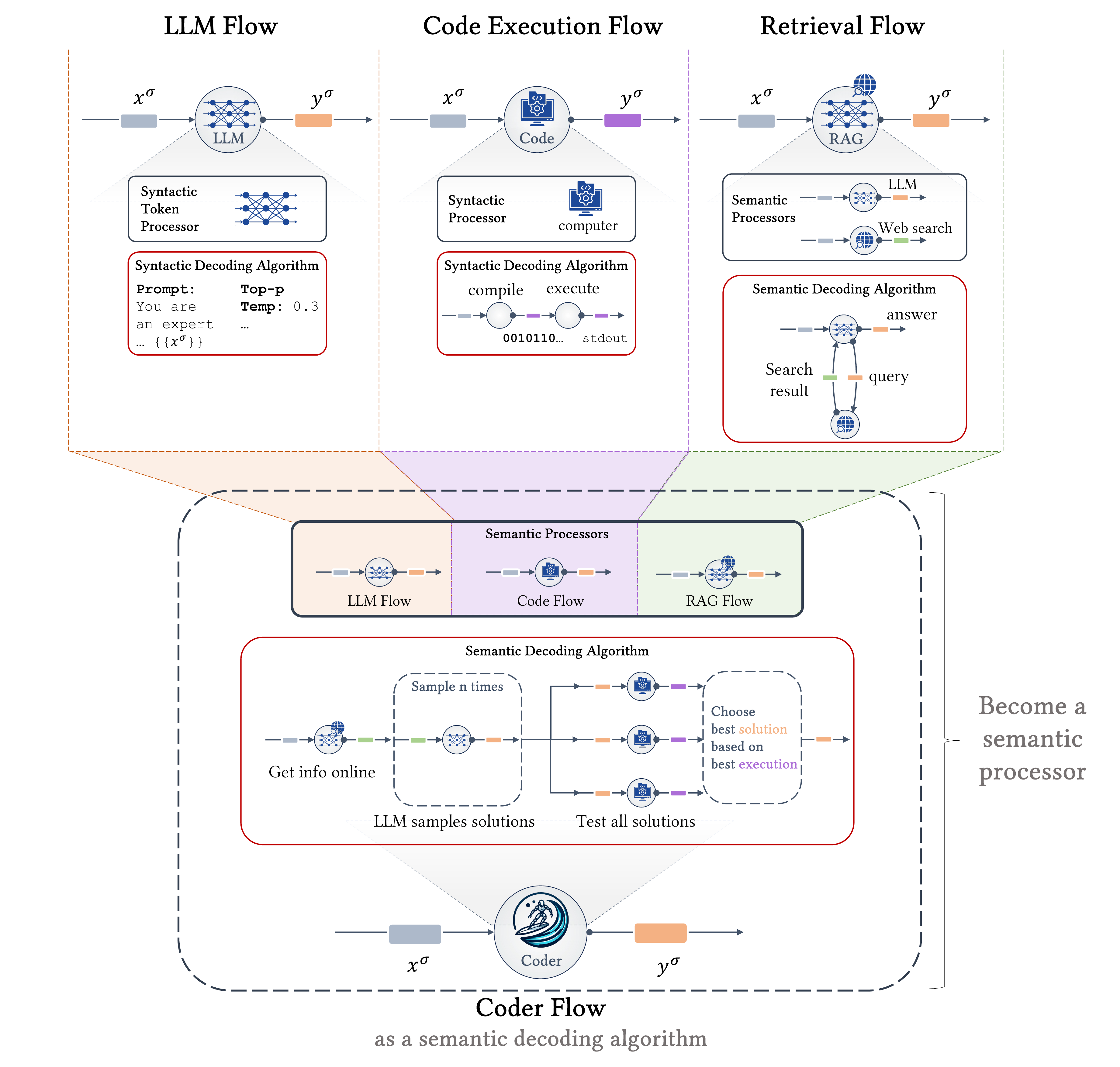}
    \caption{\textbf{Illustrating compositionality of Flows.} On the top, we see three semantic processors. The first one is implemented using a prompted language model, the second is a code executor wrapped in a Flow to enable communication, and the last one is a semantic decoding algorithm itself (Retrieval Augmented Generator), orchestrating both an LLM and a search engine. Together, these three semantic processors are orchestrated as part of a new Flow, producing a robust semantic decoding algorithm designed to solve coding problems. In this example, the Retrieval Flow initially extracts useful information from the web. Subsequently, the LLM Flow samples many candidate solutions, each of which is executed and tested by the code execution Flow. Finally, the answer to be returned is selected. This programmer Flow can then serve as a semantic processor for others to utilize. Flows, implementations of semantic decoding algorithms, become semantic processors themselves as they read and generate semantic tokens. They can then join the pool of semantic processors available for other Flows to utilize. This enables an open-ended complexity growth, with modular building blocks seamlessly interfaced through communication via semantic tokens. The Flows abstraction underscores that the distinction between semantic processors and semantic decoding algorithms is only in perspective.}
    \label{fig:flows}
\end{figure*}

\subsection{Flows as Semantic Decoding Algorithms}
The semantic decoding perspective we propose emerges from \textit{Flows} \cite{josifoski2023flows}, an abstraction for modeling structured interactions among AI systems, tools, and humans. 
This abstraction revolves around \textit{Flows} as compositional, self-contained, goal-driven entities that execute a semantically meaningful unit of work and communicate solely through semantic tokens. Flows represent semantically meaningful blocks of computation performed either by directly utilizing a ``tool,'' as in \textit{Atomic Flows}, or emerges from purposeful interactions among other Flows, as in \textit{Composite Flows}. Crucially, in the \textit{Flows} abstraction, the notion of a \textit{tool} is general and encompasses everything from a database, a compiler, or a search engine to powerful AI systems like LLaMA \cite{Touvron2023LLaMAOA}, Stable Diffusion \cite{DBLP:journals/corr/abs-2112-10752}, and GPT-4 \cite{OpenAI2023GPT4TR}; or a human. Fundamentally, the concept of an Atomic Flow lifts all tools into a single semantic space shared with Composite Flows, in which they can all communicate via semantic tokens (\ie, messages). 

Flows \textit{are} semantic decoding algorithms as they orchestrate semantic processors to produce a useful semantic token as output. The semantic decoding perspective is concerned with the optimization that Flows performs in the semantic space (the \emph{what}). Whereas the framework of \textit{Flows} is the conceptual tool enabling the design and implementation of these compositional interactions (the \emph{how}). 
Notably, a semantic decoding algorithm -- implemented by a Flow -- becomes a black-box engine that reads a semantic token as input and produces a semantic token as output, therefore becoming a semantic processor itself for other Flows to utilize. This compositional property of Flows enables open-ended complexity growth by constantly increasing the pool of existing semantic processors, blurring the frontier between a semantic decoding algorithm and a semantic processor. A Flow can be both a semantic processor within a larger computation or a semantic decoding algorithm when it is itself the outermost algorithmic layer under scrutiny for understanding what type of optimization it is performing to construct its output semantic token.
\Figref{fig:flows} illustrates the compositional properties of \textit{Flows} and its close connection to semantic decoding through an example. Throughout the rest of the paper, we use the terms Flow and semantic decoding algorithm interchangeably.

Beyond the benefits at the conceptual level, the principled abstraction of \textit{Flows} 
and its accompanying library aiFlows\footnote{\url{https://github.com/epfl-dlab/aiflows/}} enable seamless asynchronous and distributed execution, which are necessary for many promising directions, as discussed in \Secref{sec:sem_decode} and \Secref{sec:research_opportunities}. 
Furthermore, the library comes with \textit{FlowVerse}, a repository of Flows that can be readily used, extended, or composed into more complex Flows. \textit{FlowVerse} is open for anyone to contribute.

\begin{summarybox}
\textbf{Summary of the decoding perspective:} \\
A \textbf{decoding algorithm} is an algorithmic layer on top of token processors that orchestrates the computation over tokens to extract a high-utility output. 
A \textbf{semantic decoding} algorithm, in turn, is a decoding algorithm orchestrating semantic processors manipulating semantic tokens. It represents the semantic counterpart to the well-known problem of syntactic decoding from auto-regressive language models.

A semantic decoding algorithm is itself a semantic processor, ready to be employed by other, more complex, semantic decoding algorithms. This compositional property, emerging at the semantic level, enables \textbf{open-ended} complexity growth. 
The compositonal \textbf{\textit{Flows}} framework capitalizes on this property to enable and facilitate the modeling, implementation, and systematic study of arbitrarily complex structured interactions between tools, models, and humans.
Then, certain equivalences come to light: \textit{Flow} $\cong$ \textit{semantic decoding algorithm} $\cong$ \textit{semantic processor}. \textit{Flows} offers a playground for engineering semantic decoding algorithms. 
\end{summarybox}

\section{Semantic Decoding: Optimization in the Semantic Space}
\label{sec:sem_decode}
By adopting the semantic decoding perspective, we can analyze the orchestrated interactions of AI, tools, and humans based on the optimization processes they perform and the heuristics they use to navigate the semantic space. When compared to the syntactic level, the semantic level offers distinct advantages that semantic decoding algorithms can leverage. These advantages include:

\quad \textbf{Effectiveness in exploration:} 
The structure of the semantic space induced by the semantic processors inherits the intrinsic meaning associated with the tools we build and provides us value. Thus, by definition, the semantic space comes with a natural structure that makes it more meaningful for the optimization of utility. Syntactic decoding, on the other hand, consistently encounters the core issue that a semantic concept can materialize in vastly different syntactic forms, making optimization in the syntactic space inefficient and disconnected from the semantic space. 
Furthermore, unaffected by the auto-regressive nature of syntactic decoding, semantic decoding algorithms can explore the output space in a non-linear fashion. This flexibility and structure create the potential for dramatically more effective decoding algorithms.         

\quad \textbf{Human interpretability:} 
Semantic tokens inherently carry meaning, allowing humans to fully understand the computations and optimization performed by the semantic decoding algorithm. As semantic processors themselves, humans can seamlessly engage in these algorithms' computations.

\quad \textbf{Open-endedness:}
Semantic decoding algorithms operate over semantic tokens. Reading and generating semantic tokens, like other semantic processors, makes them a valid semantic processor. With the internal optimization abstracted away, semantic decoding algorithms can be readily interfaced with the ever-growing pool of semantic processors, including other decoding algorithms.
This compositional property is critical for enabling the design and implementation of processes with open-ended complexity. 



The design space for semantic decoding algorithms is vast, and we're just scratching the surface of its full potential. As a community, we still haven't properly formalized the problem and lack methods for systematically discovering, crafting, and learning semantic decoding algorithms. 
In \Secref{sec:research_opportunities}, we explore opportunities for future research. In this section, we categorize the types of search and optimization that can be performed in the semantic space, drawing connections to concepts from the world of syntactic decoding.

\subsection{Heuristic Decoding Patterns: Grammars of Thoughts}
\label{ssec:patterns}

A simple strategy for navigating the semantic space is to rely on predefined workflows dictating which types of semantic tokens should be generated at which moments. This programmed exchange of semantic tokens among semantic processors is designed to ensure a robust progression towards a high-utility final output. This line of research has been initiated by the Chain-of-Thought (CoT) method \cite{CoT}, a prompting method enforcing the generation of intermediate semantic tokens before producing a final output. 

Several works have studied more sophisticated reasoning patterns, incorporating additional semantic processors. For example, ReAct employs a two-step approach of switching between a reasoning and an acting step. This allows the model to leverage external tools to generate intermediary semantic tokens. 
Some research splits the problem-solving process into two phases implemented by dedicated planning and acting Flows \cite{wang2023describe, wang2023jarvis1, gao2023pal, paul2023refiner, josifoski2023flows}, or leverage feedback-giving Flows \cite{weng2023large, paul2023refiner, josifoski2023flows}. 
Exploring and systematically comparing diverse patterns across tasks is necessary for identifying the general principles dictating which patterns are beneficial for different models and problems that are still lacking \cite{josifoski2023flows}.

An interesting analogy can be drawn with the concept of constrained decoding in the syntactic decoding literature. 
Constrained decoding refers to methods enforcing constraints during inference time~\cite{tromble-eisner-2006-fast, geng-etal-2023-grammar}. For instance, one may want to prevent repeated n-grams \cite{stahlberg-byrne-2019-nmt, anderson-etal-2017-guided}, or enforce predefined structural constraints on the output \cite{hokamp-liu-2017-lexically, hu-etal-2019-improved, post-vilar-2018-fast, josifoski-etal-2022-genie, deutsch-etal-2019-general, shin-etal-2021-constrained}, expressible with formal grammars \cite{scholak-etal-2021-picard, roy2022benchclamp, geng-etal-2023-grammar}.

Then, it is worth contemplating what the semantic equivalent of a grammar constraint would be. Such a \textit{grammar of thoughts} would represent formal constraints on trajectories' structure in the semantic space, akin to generalized reasoning patterns. Planning before acting, validating with external tools before answering, and attempting to refute one's own answer would be simple examples of such grammar.

\subsection{Meta-Heuristics: Sampling and Value-Guided Search in Semantic Space}
\label{ssec:guided_search}

At the syntactic level, various decoding strategies exploit the probabilistic nature of language models, including top-p \cite{DBLP:conf/iclr/HoltzmanBDFC20}, top-k \cite{fan-etal-2018-hierarchical}, and stochastic beams \cite{pmlr-v97-kool19a,meister-etal-2021-conditional}, using sampling of tokens to better explore the space of possible output sequences.

Given that language models are also probabilistic generators of semantic tokens, exploration of the semantic space through sampling naturally arises. Many existing semantic decoding algorithms are built on the foundation of sampling semantic tokens. For instance, self-consistency \cite{wang2023selfconsistency} extends CoT by sampling multiple chains, thereby exploring different reasoning paths. This concept is further expanded to non-linear chains with approaches like Tree-of-Thought \cite{yao2023tree, long2023large, xie2023selfevaluation} and Graph-of-Thoughts \cite{besta2023graph, yao2023chainofthought}.

To ensure navigation towards high-utility regions of the output space, complementing sampling with an external signal, such as a value model -- an estimator of the final utility given the current trajectory -- can often be beneficial. 
At the syntactic level, this idea is embodied by Value-Guided Beam Search (VGBS) \cite{NIPS2017_2b24d495, DBLP:conf/cvpr/RenWZLL17, krishna-etal-2022-rankgen}, a variant of beam search utilizing a value model in addition to the model's likelihood to decide which next token to sample. Monte-Carlo tree search (MCTS) \cite{chaffin-etal-2022-ppl, josifoski-etal-2023-language} extends this idea by performing simulations to explore the utility of future outcomes before making the next decision.

At the semantic level, the idea of a value model can be implemented very flexibly by leveraging the rich ecosystem of semantic processors. These processors offer extensive possibilities for incorporating external signals in the computation. Language models prompted to perform different roles or with different side information can provide feedback by estimating the value of a current stream of semantic tokens \cite{xue2023rcot, shinn2023reflexion, paul2023refiner}. Tools or humans with access to external information, acting in the world, or executing code can further provide reliable grounded estimates of utility \cite{gao2023pal, chen2023program, josifoski2023flows, li2023chain}. Already, \citet{ding2023thoughts} hint at the idea of performing MCTS-based planning in semantic space.

Semantic decoding algorithms like FunSearch \cite{FunSearch2023} and PromptBreeder \cite{fernando2023promptbreeder} utilize a language model as the sampler of candidate solutions (semantic tokens) within evolutionary algorithms optimizing a population of candidate solutions. The fitness function can be viewed as a simple value model guiding the decoding of the next semantic tokens, i.e., the next candidate solutions. These examples showcase the effectiveness of combining optimization through search based on sampling from a competent sampler, i.e., a language model.

Currently in its early stages, this research direction lacks proper formalization within the optimization domain. Nevertheless, it shows great potential to enable the emergence of new types of AI capabilities, as evidenced by systems such as FunSearch~\cite{FunSearch2023}.

\subsection{Learning the Flow: Embracing Optimization in Semantic Space}
\label{ssec:learning_flow}

Moving beyond heuristics and meta-heuristics, one could fully commit to the optimization perspective within the semantic space.
At a micro level, optimization algorithms could focus on enhancing performance by training the semantic processors to collaborate more effectively. We term this approach as \textit{learning to collaborate}. Alternatively, taking a more holistic view of the problem, optimization might involve training a controller that determines which semantic processor to invoke at each time step along with the appropriate parameters. We refer to this approach as \textit{learning to orchestrate}. For example, one could utilize reinforcement learning \cite{BahdanauBXGLPCB16} or reward-based supervised learning \cite{NIPS2016_2f885d0f, rafailov2023direct, DBLP:journals/corr/abs-2308-08998} to backpropagate a learning signal through semantic tokens, optimizing over sequences of syntactic tokens directly. Systems such as AutoGPT \cite{autogpt2023} or BabyAGI \cite{babyagi2023} already use a prompted language model as a heuristically optimized controller of a general-purpose semantic decoding algorithm. Moreover, learning to orchestrate can be seen as a planning problem, and therefore, ideas from the extensive literature on planning could be explored in this context.

This general notion of learning the semantic decoding algorithm builds upon previous work at the syntactic level. For instance, \citet{wiseman-rush-2016-sequence} introduced a differentiable relaxation of beam-search, and \citet{pmlr-v97-collobert19a} developed a fully differentiable beam-search that can be optimized during training.

\begin{summarybox}
\textbf{Summary of semantic decoding optimization:} \\
Decoding directly in the semantic space offers many benefits, including flexible and effective exploration of the meaningfully structured semantic space, human-interpretable computation, and open-endedness enabled by the compositional nature of Flows.

The optimization performed in the semantic space can be broadly categorized into three main types:
(i) \textbf{Heuristic decoding patterns}: programmed interactions such as CoT, ReAct, or meta-patterns like planning before acting and iterative, feedback-based refinement.
(ii) \textbf{Sampling and value-guided search}: interactions defining optimization strategies that explore the semantic space by strategically sampling semantic tokens and leveraging a value function to guide the process.
(iii) \textbf{Learning to optimize in the semantic space}: \textit{learning to decode} by training the semantic decoding algorithms and their components. For instance, learning to collaborate, orchestrate, search, learn effective reasoning patterns, and more.
\end{summarybox}

\section{Research and Application Opportunities}
\label{sec:research_opportunities}
In the previous section, we highlighted the distinctive benefits of the semantic decoding perspective and outlined what kind of optimization and search can be performed in semantic space. In this section, we shift towards applications and research questions opened up by the semantic decoding perspective. While not exhaustive, this section serves to illustrate the depth and potential of adopting the semantic decoding viewpoint.

\subsection{Prompt Engineering and Meta-Prompt Engineering}
The ability of LLMs to adapt to the information in their context has led to remarkable breakthroughs across fields \cite{NEURIPS2020_1457c0d6, NEURIPS2022_8bb0d291}, with prompt engineering playing a central role in these advancements. For example, compared to traditional prompting methods, CoT \cite{CoT} has demonstrated a threefold increase in performance on the GSM8K dataset. Likewise, for competitive coding, a fixed collaboration pattern between two GPT-4 instances -- a coder and a critic with access to a code executor -- increases the solve rate by 1.8 times \citep{josifoski2023flows}. These represent just a glimpse of the many examples where LLM performance has been significantly improved without fine-tuning. Researchers continue to push boundaries by crafting complex patterns to tackle increasingly complex tasks \cite{FunSearch2023} and exploring meta-prompting methods \cite{fernando2023promptbreeder}. Viewing (meta-)prompt engineering as optimization in semantic space opens the door to novel ideas and provides a principled structure for the research efforts in this particularly active and promising direction.

\subsection{Synthetic Data Flow}
Data, itself composed of semantic tokens, can become the focal point for Flows dedicated to manipulating or synthesizing data.
Similarly, specialized trainer Flows can be aimed at training semantic processors (e.g., models) or even entire \mbox{(sub-)Flows}. 
By combining trainer Flows with synthetic data generation Flows, opportunities emerge for creating sophisticated self-training loops. 
The synthetic data generation Flows can leverage domain knowledge \cite{tang2023does}, task properties \cite{lu2024mathgenie,veselovsky2023generating,josifoski-etal-2023-exploiting}, or collaboration \cite{abdullin2024synthetic}, and synthesize data of notably higher quality than what a single model or simple heuristics can achieve. 
This sets the stage for effective self-improvement loops where a language model participates in a semantic decoding algorithm producing high-quality synthetic data. 
Then, the language model improves itself through fine-tuning, thereby improving the Flow's capacity to generate even better synthetic data in a virtuous cycle \cite{silver2017mastering,burns2023weaktostrong,singh2023human,chen2024selfplay}.
An example of such a Flow is MAGDi \cite{chen2024magdi}, a framework designed to distill reasoning interactions among multiple LLMs into smaller ones. This approach surpasses single-teacher distillation \cite{li-etal-2023-symbolic, magister-etal-2023-teaching} and finetuning based on reasoning trajectories sampled from GPT-4 \cite{chen2023fireact}.

\subsection{Human in the Loop and Human-Computer Interaction}
Humans operate within the semantic space and can seamlessly integrate into semantic decoding algorithms as just another type of semantic processor. 
This integration creates numerous opportunities to explore diverse human-computer interactions within a principled optimization-based framework. 
This framework can leverage human cognition as a computational input for semantic decoding algorithms aimed at maximizing a utility function. For instance, humans can provide nuanced low-level feedback, offer high-level guidance, or simply observe the ongoing exchange of semantic tokens \cite{josifoski2023flows,cai2023humanintheloop,xiao2023llm,kim2024meganno,li2023collaborative}. Moreover, humans can dynamically switch roles during the execution, intervening, pausing the process, and resuming as necessary. 
Flow engineering methods, such as searching for heuristic reasoning patterns (\Secref{ssec:patterns}), crafting guided search methods (\Secref{ssec:guided_search}), or learning the Flow (\Secref{ssec:learning_flow}), can be designed to optimize objectives that involve maximizing a utility function while minimizing the cognitive cost for the human in the loop \cite{horvitz1999principles}.

\subsection{General AI Assistants}
Considerable research efforts are currently focused on constructing general-purpose assistants. Existing methodologies involve utilizing LLMs as controllers, together with selected tools to enhance \textit{cognitive} abilities \cite{babyagi2023,autogpt2023,wang2023jarvis1,wu2024oscopilot}. However, recent evaluations on the GAIA benchmark \cite{mialon2023gaia} have revealed a notable performance gap compared to humans, highlighting the need for improvements. For example, GPT-4, when equipped with tools, achieves only 15\% accuracy, while humans attain 92\% accuracy.
Examining this challenge through the perspective of semantic decoding, it can be framed as the development or learning of a general-purpose semantic decoding algorithm.

\subsection{Evaluation and Diagnostic}
Evaluating semantic decoding algorithms presents a significant challenge because they evade standard controlled benchmarks for two primary reasons. Firstly, the issue of data contamination arises because language models are trained on vast amounts of internet data, potentially including benchmark samples. To address this issue, research is needed to untangle the impact of memorization and generalization across various contexts\cite{josifoski2023flows,shi2023detecting,golchin2024time}.

Secondly, the dynamic nature of the environment leads to the evolution of AI systems over time. For instance, a search engine's results can vary not only due to algorithmic changes but also due to updates in the real world. Overcoming this challenge necessitates advancing diachronic evaluation methodologies specifically tailored to assess dynamic AI systems in evolving environments, as exemplified by dynamically evolving benchmarks \cite{li2023latesteval}.

Despite these evaluation challenges, it remains imperative to develop methods that effectively discern between effective and ineffective semantic decoding algorithms. This understanding is crucial for informing users about the expected behavior of AI systems and enabling practitioners to enhance system performance. Tools from causality, such as causal mediation analysis, can be particularly useful in recognizing which components or messages were critical to the computation's success or failure. These diagnostic considerations are tightly linked to the interpretability challenges discussed below.

\subsection{Interpretability, Control, and Error Mitigation}

\xhdr{Interpretability}
A semantic decoding algorithm, as any AI system, is subject to questions of explainability to understand \textit{why} some outputs were produced \cite{Woodward2003-WOOMTH,Potochnik2017-POTIAT-3,lipton_2018}. One approach is to perform behavioral analyses by manipulating inputs and observing the resulting effects on the outputs. Additionally, model-agnostic feature importance methods, such as LIME \cite{Ribeiro_lime} or SHAP \cite{NEURIPS2020_ecb287ff}, can be expanded to include semantic tokens rather than just syntactic features. 
Opening the black box to inspect the network of semantic tokens exchange leads the path toward mechanistic interpretability of semantic decoding algorithms. This involves manipulating intermediate semantic tokens, placing the system in a counterfactual state, and then resuming the computation to precisely measure the impact of each computational step \cite{PearlMackenzie18,Geiger-etal:2022:SAIL}. In general, interpreting semantic decoding algorithms is somewhat simpler than language models because semantic tokens serve as discrete, semantically meaningful bottlenecks in the computation that can be easily inspected and intervened upon.

\xhdr{Control and ethics}
Closely linked to the issue of explainability is the notion of control \cite{Woodward2003-WOOMTH,PearlMackenzie18}. By breaking down computation into discrete, human-interpretable chunks, we obtain many levers for controlling the computation. 
This can be done manually by humans or automatically by appending dedicated semantic processors to correct, adjust, or prohibit some predefined undesirable intermediate steps, thus preventing undesired outcomes. 
For example, specialized \textit{ethics semantic processors} (possibly complex flows themselves) can be inserted at critical steps to exclude unwanted semantic tokens and steer the computation toward areas of the output space that align with predefined ethical objectives \cite{gallegos2024selfdebiasing}. Such ethics components might leverage expert debiasing methods that scrutinize semantic tokens for potential societal biases and either remove them by reformulating or providing informed feedback.

\subsection{Richer Semantic Spaces}
\xhdr{From syntactic to semantic and back} 
One advantage of semantic tokens is that they are readily understood by humans, rendering semantic decoding algorithms inspectable, interpretable, and conducive to human participation. Yet, the decoding perspective does not inherently require semantic tokens to be human-interpretable. One could imagine a collaboration between LLMs and tools that, for the sake of communication efficiency or effectiveness, relies on new made-up languages. These new semantic spaces can be learned as part of the \textit{learning the flow} pipelines to optimize the amount of useful information exchanged per message.

\xhdr{Multimodal semantic tokens}
Another way in which the concept of semantic token can be stretched is by considering other modalities.
The semantic decoding perspective easily accommodates more general informational units \cite{reed2022generalist}.
Developing more competent multimodal language models is an active area of research \cite{wu2023multimodal}. Multimodal models can easily participate in semantic decoding algorithms, significantly enlarging the semantic space and the variety of signals available to guide exploration toward high-utility outputs.

\subsection{Infrastructure}
In order to support the innovations discussed above, it is imperative to have robust infrastructures. 
Significant efforts have already been invested in creating efficient abstractions \cite{Lu2023ChameleonPC,li2023camel,Shen2023HuggingGPTSA,langchain2022,DBLP:journals/corr/abs-2308-00352,DBLP:journals/corr/abs-2308-08155}. However, the \textit{Flows} framework \cite{josifoski2023flows} stands out by introducing the modularity and compositionality essential for systematically constructing intricate systems (\Secref{sec:decoding}). Its support for concurrent and distributed execution unlocks creative applications like FunSearch \cite{FunSearch2023}, PromptBreeder \cite{fernando2023promptbreeder}, meta-reasoning Flows \cite{josifoski2023flows}, and arbitrary peer-to-peer collaborations between Flows, which could define anything from a stateless tool to a general assistant or autonomous agent.

Beyond proper abstractions, \textit{Flows} and semantic decoding algorithms enable a level of complexity and flexibility that comes with additional technical challenges. For instance, \textit{Flows} supports open-ended complexity and allows practitioners to freely develop and publicly deploy Flows, thereby necessitating systematic and contextual Flow indexing and retrieval (\ie, a search engine over Flows). Additionally, executing Flows may depend on resource-intensive semantic processors, such as commercial LLMs, which necessitates efficiency optimizations to minimize resource usage or latency while maintaining utility.

\xhdr{Flow indexing and retrieval}
Similar to how a web page provides information or access to some services, a Flow provides a meaningful computation. Given the ability to publicly deploy Flows and the infrastructure for peer-to-peer communication between Flows, the need for a search engine over the space of Flows is going to grow as the number of publicly accessible Flows increases. Developing scalable yet effective methods for indexing Flows (\ie, semantic computation), akin to PageRank for web pages, is bound to become an important research direction in the near future.

\xhdr{Efficiency optimization}
Improving the efficiency of a Flow can be achieved by enhancing the efficiency of its semantic processors. Thankfully, a rich body of work is dedicated to improving the efficiency of language models' inference. Techniques like batching and key-value caching can mitigate the cost and latency of decoding long sequences autoregressively \cite{pope2022efficiently}. Speculative decoding methods, which involve generating candidate samples using smaller models and then refining or filtering these samples with larger models, represent a different approach to optimization \cite{NEURIPS2022_6fac9e31, chen2023accelerating, kim2023speculative, abs-1811-03115}. Further optimizations include memory usage reduction through weight quantization \cite{yao2023zeroquantv2, dettmers2022llmint8, frantar2023gptq, dettmers2023qlora, xiao2023smoothquant, ashkboos2023quik}, and leveraging non-homogeneous computational costs in the transformer graph, as exemplified by SkipDecode \cite{delcorro2023skipdecode} and PowerInfer \cite{song2023powerinfer}.

At the semantic level, caching Flow calls with soft, approximate caches \cite{ramírez2023cache} and dynamically routing queries to different models based on their properties \cite{hu2023enabling, liu2023dynamic, šakota2024flyswat, yue2023large, lee2023orchestrallm} are feasible strategies. Concurrent execution of semantic processors, such as Skeleton-of-Thoughts \cite{ning2023skeletonofthought}, also contributes to latency gains. Despite these promising approaches, the concept of \emph{speculative semantic decoding} remains relatively unexplored. Systematic studies could explore replacing expensive sub-flow components with faster, more economical modules trained to emulate or cache computations from costly sub-Flows.

\section{Conclusion}
\label{sec:discussion}
This paper proposes semantic decoding, which formalizes LLMs, humans, and other tools as semantic processors that read and generate semantic tokens, and the collaborations between them as optimization processes in the semantic space. These optimization processes correspond to semantic decoding algorithms and aim to find high-utility semantic tokens as answers to queries. This perspective allows us to systematically study the design space of semantic decoding algorithms based on the optimization they perform.

The transition from syntactic to semantic tokens represents a pivotal moment in the development of AI systems. By abstracting the syntactic details, we can conceptualize LLMs, humans, and tools as a form of computation that operates directly within the space of meaningful concepts. The potential offered by orchestrated collaborations between them is vast, yet we are only scratching the surface of what is possible.

We believe that the concept of Flows \cite{josifoski2023flows}, supported by the aiFlows library\footnote{\url{https://github.com/epfl-dlab/aiflows}}, serves as a practical framework for engineering robust, modular, and compositional semantic decoding algorithms.

\section*{Acknowledgments}
West's lab is partly supported by grants from
Swiss National Science Foundation (200021\_185043, TMSGI2\_211379),
Swiss Data Science Center (P22\_08),
H2020 (952215), Microsoft Swiss Joint Research Center, and Google, and by generous gifts from Facebook, Google, and Microsoft.

\clearpage

\section*{Glossary of Important Terms}
\label{sec:glossary}
\begin{center}
\begin{description}
\item[Syntactic Token]
Basic representational units of language, typically defined by the tokenizer in modern NLP systems.

\item[Semantic Token]
Semantically coherent units of text, also known as \textit{thoughts}, representing meaningful concepts or ideas.

\item[Syntactic Processor]
Entities that process and manipulate syntactic tokens, such as parsers or language models.

\item[Semantic Processor]
Entities that process and manipulate semantic tokens, including LLM-based systems, humans, and various tools such as search engines and code executors. Notably, a semantic processor can be implemented by a semantic decoding algorithm itself, showcasing the recursive compositional property that emerges at the semantic level.

\item[Utility Function]
A function that scores candidate solutions, assessing their effectiveness in solving a task for a given input query. The goal of an AI system is to produce an output with high utility.

\item[Value Model]
A model that estimates the expected utility of partial trajectories during a decoding process, guiding decoding strategies towards regions of expected high utility.

\item[Decoding]
The process of extracting high-utility sequences or trajectories, which can be performed at both the syntactic and semantic levels.

\item[Syntactic Decoding]
The process of extracting high-utility sequences of syntactic tokens, often relying on sampling tokens and guided by value models or heuristics. Examples include top-p, value-guided beam search, and Monte Carlo tree search decoding.

\item[Semantic Decoding]
The process of extracting high-utility trajectories in semantic space, leveraging sampling, semantic-level value models, or heuristics to optimize the output. Examples include methods that orchestrate interactions between semantic processors, such as chain-of-thought, tree-of-thoughts, ReAct, and FunSearch. When ignoring computational details, a semantic decoding algorithm becomes a semantic processor, showcasing the recursive compositional property that emerges at the semantic level.

\item[Flows]
An abstract model of communication between LLMs, humans, and tools that supports the implementation of modular and compositional semantic decoding algorithms. Flows \textit{are} semantic decoding algorithms in that they specify a general framework to implement them. We use the term \textit{semantic decoding algorithm} to emphasize the optimization aspect in semantic space, while we use the term \textit{Flows} to focus on the engineering and implementation aspects.
\end{description}
\end{center}

\clearpage
\bibliography{anthology,main}
\bibliographystyle{acl_natbib}

%



\end{document}